\documentclass[runningheads]{llncs}
\usepackage{graphicx}

\usepackage{booktabs} 

\usepackage{booktabs} 
\makeatletter  
\newif\if@restonecol  
\makeatother

\usepackage[linesnumbered,ruled,vlined]{algorithm2e}
\usepackage{algpseudocode}  
\usepackage{amsmath}
\usepackage{amsfonts}

\makeatletter
\newcommand{\HEADER}[1]{\ALC\underline{\textsc{#1}}\begin{ALC@g}}
\newcommand{\ENDHEADER}{\end{ALC@g}}
\makeatother

\usepackage[pagebackref,breaklinks,colorlinks]{hyperref}

\begin{document}
\title{Fine-Grained Image Style Transfer \\with Visual Transformers}
%
%
\author{Jianbo~Wang\inst{1} \and
        Huan~Yang\inst{2} \and
        Jianlong~Fu\inst{2} \and \\
        Toshihiko~Yamasaki\inst{1} \and
        Baining~Guo\inst{2}
}
\authorrunning{J. Wang et al.}
%
\institute{The Univerisity of Tokyo \email{\{jianbowang815,yamasaki\}@cvm.t.u-tokyo.ac.jp}
\and Microsoft Research \email{\{huayan,jianf,bainguo\}@microsoft.com}
}

\maketitle              
\begin{abstract}
With the development of the convolutional neural network, image style transfer has drawn increasing attention. However, most existing approaches adopt a global feature transformation to transfer style patterns into content images (e.g., AdaIN and WCT). Such a design usually destroys the spatial information of the input images and fails to transfer fine-grained style patterns into style transfer results. To solve this problem, we propose a novel \textbf{ST}yle \textbf{TR}ansformer (\textbf{STTR}) network which breaks both content and style images into visual tokens to achieve a fine-grained style transformation. Specifically, two attention mechanisms are adopted in our STTR. We first propose to use self-attention to encode content and style tokens such that similar tokens can be grouped and learned together. We then adopt cross-attention between content and style tokens that encourages fine-grained style transformations. To compare STTR with existing approaches, we conduct user studies on Amazon Mechanical Turk (AMT), which are carried out with 50 human subjects with 1,000 votes in total. Extensive evaluations demonstrate the effectiveness and efficiency of the proposed STTR in generating visually pleasing style transfer results\footnote{Code is available at \href{https://github.com/researchmm/STTR}{https://github.com/researchmm/STTR}.}.
\end{abstract}

\section{Introduction}
Image style transfer has been receiving increasing attention in the creation of artistic images. Given one content and one style reference image, the model will produce an output image that retains the core elements of the content image but appears to be ``painted'' in the style of the reference image. It has many industrial applications, for example, clothes design~\cite{clothes}, photo and video editing~\cite{video,zhang2013style,liu2017photo,virtusio2021neural}, material changing~\cite{material}, fashion style transfer~\cite{kim2019style,liu2019swapgan}, virtual reality~\cite{vr}, and so on.

In recent years, style transfer employs deep neural networks. Those methods could be divided into three categories: 1)~optimization-based methods, 2)~feed-forward approximation, and 3)~zero-shot style transfer. Gatys et al.~\cite{gatys2016} propose to optimize the pixel values of a given content image by jointly minimizing the feature reconstruction loss and style loss. It could produce impressive results, but it requires optimizing the content image for many iterations for any content-style image pairs which are computationally expensive. 
To solve this problem, other researchers addressed this problem by building a feed-forward network~\cite{johnson2016perceptual,li2016precomputed,ulyanov2016texture,dumoulin2016learned,chen2017stylebank,li2017diversified,zhang2018multi}, to explicitly learn the mapping from a photograph to a stylized image with a particular painting style. Thus it requires retraining a new model for any unseen new styles.
 Zero-shot style transfer is more effective since it could handle various styles and transfer images given even unseen styles. Huang et al.~\cite{adain} propose an arbitrary style transfer method by matching the mean-variance statistics between content and style features, usually named adaptive instance normalization (AdaIN).  AdaIN first normalizes the input content image, then scales and shifts it according to parameters calculated by different style images. A more recent work replaces the AdaIN layer with a pair of whitening and coloring transformations~\cite{wct}. Many works follow the formulation of AdaIN and further improve it~\cite{kitov2019depth,aast}.

However, the common limitation of these methods is that simply adjusting the mean and variance of feature statistics makes it hard to synthesize complicated style patterns with rich details and local structures. As shown later in Fig.~\ref{fig:comparison}, Gatys et al.'s method~\cite{gatys2016}, AdaIN~\cite{adain}, and WCT~\cite{wct} always bring distorted style patterns in the transferred results, which makes it hard to recognize the original object from the style transferred image. In a more recent work~\cite{stytr}, Deng et al. propose the StyTr$^{2}$ to learn semantic correlations between the content and style features by attention. Although StyTr$^{2}$ could produce visually pleasing results, there still exists structure distortion as shown later in Fig.~\ref{fig:comparison}. This is because StyTr$^{2}$ adopts a shallow feature extractor without pre-trained weights which limits its abilities to capture and transfer style patterns and makes foreground and background objects indistinguishable in the resulting image. Thus, how to learn a better representation to assemble the content and style patterns and transfer fine-grained style patterns while keeping content structure is still a challenging problem.

In this paper, we adopt a Transformer to tackle this problem. 
Recently, the breakthroughs of Transformer networks~\cite{transformer} in the natural language processing (NLP) domain has sparked great interest in the computer vision community to adapt these models for vision tasks. 
Inspired by those works which explicitly model long-range dependencies in language sequences and images, we develop a Transformer-based network to first break content and style images into visual tokens and then learn the global context between them.
As similar content tokens lean to match with the same style tokens, fine-grained style transformation can be obtained. Specifically, the proposed STTR mainly consists of four parts: two backbones to extract and downsample features from the inputs (i.e., content and style images), two self-attention modules to summarize the style or content features, a cross-attention module to match style patterns into content patches adaptively, and a CNN-based decoder to upsample the output of cross-attention module to reconstruct the final result. We implement this by Transformer since its encoder consists of self-attention module while the decoder has cross-attention module to compute the correlations between content and style tokens.

Two backbones first extract compact feature representation from given content and style images and downsample them to a smaller spatial size. This could help to reduce the computation cost in the Transformer. Secondly, the self-attention modules encourage the style (or content) features with similar structures and semantic meanings to group together.
Different from previous work, feeding content and style into the self-attention model before matching them could help to separate them into different distinctive semantic vectors (see Sec.~\ref{sec:ablation-encoder} for more information). The cross-attention module in the decoder further incorporates style into content harmoniously without hurting the content structure. After that, a CNN-based decoder upsamples the output of the decoder to obtain the stylized results. The details of network architecture could be found in Sec.~\ref{sec:approach}.

Our main contributions are as follows. 

\begin{itemize}
\item We propose a fine-grained Transformer-based image style transfer method, 
capable of computing very local mapping between the content and style tokens. 

\item To better understand the effectiveness of STTR to model global context between style and content features, we provide a detailed ablation study to explore the contribution of the Transformer architecture and losses.
\item We evaluated the performance of the proposed STTR with several state-of-the-art style transfer methods. Experimental results show that the STTR has a great capability of producing both perceptual quality and details in transferring style into a photograph. We also provide frame-wise video style transfer results which further demonstrate the effectiveness of our approach. 
\end{itemize}

\begin{figure*}[t]
  \centering
  \includegraphics[width=1.01\linewidth]{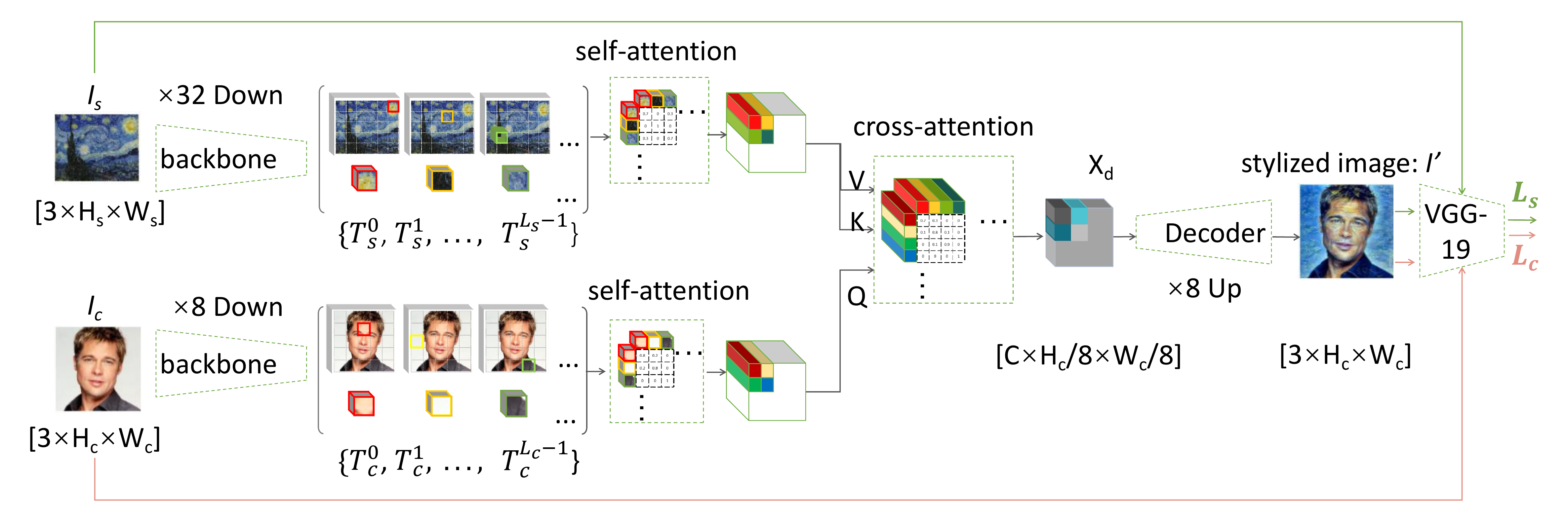}
  \caption{An overview of our proposed Style Transformer (STTR). Our framework consists of four parts: two backbones which are used to extract features from the style and content image and break them into tokens, two self-attention modules to encourage the style (or content) tokens with similar structures and semantic meanings to group together, a cross-attention module observes the outputs and adaptively match the style code to the content features, and the CNN-based decoder upsamples the output of the Transformer decoder to reconstruct the final result. We use a fixed-weight VGG-19 network to compute the content and style loss as described in Sec.~\ref{sec:loss}.
 }
  \label{fig:framework}

\end{figure*}

\section{Related work}
\subsection{Image Style Transfer}
Image style transfer is a technique that aims to apply the style from one image to another content image. Typically, neural style transfer techniques could be divided into three aspects: 1)~optimization-based methods, 2)~feed-forward approximation, and 3)~zero-shot style transfer. 

\textbf{Optimization-based methods.}
Gatys et al.~\cite{gatys2016} 
are the first to formulate style transfer as the matching of multi-level deep features extracted from a pre-trained deep neural network. The authors achieve style transfer by jointly minimizing the feature loss~\cite{mahendran2015understanding} and the style loss formulated as the difference of Gram matrices. Optimizing an image is computationally expensive and contains no learned representations for the artistic style, which is inefficient for real-time applications.

\textbf{Feed-forward approximation.}
Fast feed-forward approaches~\cite{johnson2016perceptual,li2016precomputed,ulyanov2016texture} address this problem by building a feed-forward neural network, i.e., the style transfer network, to explicitly learn the transformation from a photograph to a particular painting style. They minimize the same feature reconstruction loss and style reconstruction loss as~\cite{gatys2015neural}. However, they have to train individual networks for every style, which is very inefficient. 

\textbf{Zero-shot style transfer.} Chen et al.~\cite{chen2016fast} propose to match each content patch to the most similar style patch and swapped them. 
For fast arbitrary style transfer, Huang et al.~\cite{adain} employ adaptive instance normalization (AdaIN) to normalize activation channels for stylization. Specifically, it follows an encoder-AdaIN-decoder formulation. The encoder uses the first few layers of a fixed VGG-19 network to encode the content and style images. The AdaIN layer is used to perform style transfer in the feature space. Then a decoder is trained to invert the AdaIN output to the image spaces. Unlike AdaIN, Whitening and Coloring Transform (WCT)~\cite{wct} directly uses different layers of the VGG network as the encoders, transfers multi-level style patterns by recursively applying whitening and coloring transformation, and trains the decoders to invert the feature into an image. Many works follow the formulation of AdaIN and further improve it~\cite{kitov2019depth,aast,artflow,mccnet}. In SANET~\cite{sanet} and AdaAttN~\cite{adaattn}, the authors propose to learn semantic correlations between the content and style features by a learnable non-local module. However, those feature transformation methods always fail to maintain content structures since they simply transfer features across all spatial locations for each channel. Hong et al.~\cite{hong2021domain} propose a unified architecture for both artistic and photo-realistic stylizations. In StyTr$^{2}$, the authors propose a transformer-based style transfer model and design the tokenizer with a single convolutional layer (an unfold operation followed by a convolutional filter). Due to the shallow feature extractor for the tokenizer, the learned correspondences between the content and style tokens are random.

\begin{algorithm}[t]   
  \caption{The training process of the proposed STTR}
  \label{alg:pip}
  \KwIn{ Content Image $I_c$, Style Image $I_s$}  
  \KwOut{Stylized image $I'$}  
  $X_c \leftarrow$  extracted feature maps from $I_c$ by content backbone.\\
  $X_s \leftarrow$ extracted feature maps from $I_s$ by style backbone.\\
  Linear flat $X_c$ into $L_c$ visual tokens: $\{T_c^0,T_c^1,...,T_c^{L_c-1}\}$. 
  Linear flat $X_s$ into $L_s$ visual tokens: $\{T_s^0,T_s^1,...,T_s^{L_s-1}\}$. \\
  Compute style codes $\{C_s^0,C_s^1,...,C_s^{L_s-1}\}$ with $\{T_s^0,T_s^1,...,T_s^{L_s-1}\}$ by Transformer encoder.\\
  Match $\{T_c^0,T_c^1,...,T_c^{L_c-1}\}$ with semantic similar style codes $\{C_s^0,C_s^1,...,C_s^{L_s-1}\}$ by Transformer decoder, generate decoded visual tokens: $\{T_d^0,T_d^1,...,T_d^{L_c-1}\}$.\\
  Reshape $\{T_d^0,T_d^1,...,T_d^{L_c-1}\}$ with the same size of $X_c$, obtain $X_d$.\\
  Reconstruct $X_d$ by CNN-based decoder, obtain the final output $I'$, it has the same size with $I_c$.\\
  Compute the content and style loss defined in Sec.~\ref{sec:loss} by a fixed weight VGG-19 network.\\
  Updates the weights by minimizing the loss value.
  
\end{algorithm}  

\subsection{Visual Transformer}
Recently, there is impressive progress in the field of NLP, driven by the innovative architecture called Transformer~\cite{transformer}. In response, several attempts have been made to incorporate self-attention constructions into vision models. Those methods could be divided into two aspects: high-level vision and low-level vision. 

\textbf{High-level vision Transformer.} Transformer can be applied to many high-level tasks like image classification. 
Dosovitskiy et al. propose a Vision Transformer (ViT)~\cite{vit}, dividing an image into patches and feeding these patches (i.e., tokens) into a standard Transformer. Recently, Transformers have also shown strong performance on other high-level tasks, including detection~\cite{detr,beal2020toward,pan20203d,yuan2020temporal,zhu2020deformable}, 
segmentation~\cite{wang2020max,wang2020end,zheng2020rethinking}, and pose estimation~\cite{huang2020hand,huang2020hot,lin2020end,yang2020transpose}.

\textbf{Low-level vision Transformer.}
Low-level vision tasks often take images as outputs (e.g., high-resolution or denoised images), which is more challenging than high-level vision tasks such as classification, segmentation, and detection, whose outputs are labels or boxes. iGPT~\cite{igpt} trains GPT v2 model~\cite{gptv2} on flattened image sequences (1D pixel arrays) and shows that it can generate plausible image outputs without any external supervision. Zeng et al.~\cite{zeng2021improving} propose a Transformer-based Generation. Nowadays, researchers achieve good performance on Transformer networks for super-resolution and frame interpolation~\cite{ttsr,liu2022learning,qiu2022learning,liu2022ttvfi}. 

Image style transfer is also a sub-task of low-level vision tasks. Specifically, there exists a large appearance gap between input and output domains. This makes it difficult to produce a visually pleasing target output.

\vspace{+10pt}
\section{Approach}\label{sec:approach}
We show the pipeline of our proposed STTR in Fig.~\ref{fig:framework}. Our method takes a content-style image pair as input and produces stylized results as output. In this section, we present the details of the network architecture and loss functions accordingly. The whole training process is illustrated in Alg.~\ref{alg:pip}.

\subsection{Network Architecture}\label{sec:net-arch}

As shown in Fig.~\ref{fig:framework}, our model contains four main components. 

We will describe the network architecture as below: two CNN-based backbones to extract compact feature representations, two self-attention modules to encode the style or content features, a cross-attention module to match style patterns into content patches adaptively, and a CNN decoder to transform the combined output features from Transformer to the final output $I'$.

\subsubsection{Tokenizer} \label{sec:tokenizer}

In our model, one image is divided into a set of visual tokens. Thus, we have to first convert the input image into a set of visual tokens. We assume that each of them represents a semantic concept in the image. We then feed these tokens to a Transformer.
Let us denote the input feature map by $X\in \mathbb{R}^{H \times W \times C}$ (height $H$, width $W$, and channels $C$) and visual tokens by $T\in \mathbb{R}^{L \times C}$ where $L$ indicates the number of tokens.

Filter-based tokenizer utilizes a deep CNN-based feature extractor to gradually downsample input images to obtain feature maps. 
After that, each position on the feature map represents a region in the input space (i.e., the receptive field). If the output size of feature maps is $ H/32 \times W/32 \times C_2$ (height $H$, width $W$, and channels $C_2$), we then obtain $(H/32  \times W/32)$ patches (i.e., tokens). For each token, its dimension is $C_2$. As CNN densely slides the filter over the input image, it could produce much more smooth features. 

We further provide a detailed illustration of different tokenizers in the supplementary.

\subsubsection{Backbone}

To achieve fine-grained style transformation, inspired by the original Transformer~\cite{transformer}, one image could be divided into a set of visual tokens as one sentence consisting of a few words. Thus, we have to first convert the input image into a set of visual tokens. We assume that each of them represents a semantic concept in the image. We then feed these tokens to a Transformer.
Let us denote the input feature map by $X\in \mathbb{R}^{H \times W \times C}$ (height $H$, width $W$, and channels $C$) and visual tokens by $T\in \mathbb{R}^{L \times C}$ where $L$ indicates the number of tokens.

We adopt ResNet-50~\cite{resnet} as our backbone since it gradually downsamples the features from the input into a smaller spatial size. 
The original ResNet-50~\cite{resnet} has four layers and downsamples features by 4, 8, 16, and 32 times. The output channel of each layer is 256, 512, 1024, and 2048, respectively. For extracting style features, we choose the output from the $4{th}$ layer while for content features we choose the $2{nd}$ layer. We would like to utilize the ability of the shallow network (the $2{nd}$ layer) to extract and retain detailed information of the spatial structure. For style features, higher-level semantic features are extracted by a deeper network (the $4{th}$ layer).

In summary, starting from the input content image $I_c \in \mathbb{R}^{H_c \times W_c \times 3}$ and style image $I_s \in \mathbb{R}^{H_s \times W_s \times 3}$, two backbones will extract features independently with the size of $H_c/8 \times W_c/8 \times 512$ and $H_s/32 \times W_s/32 \times 2048$ for content and style, respectively. 

The extracted style and content features have a different number of channels as we described above. Thus we first use a $1 \times 1$ convolution to make them have the same channel number $d$. Specifically, we set $d=256$. Next, we flatten the 3D features in spatial dimensions, resulting in 2D features, $(H_c/8 \cdot W_c / 8) \times 256$ for content features while $(H_s/32 \cdot W_s/32) \times 256$ for style features.

\subsubsection{Attention Layer}
Our proposed STTR consists of an encoder module and a decoder module with several encoder or decoder layers of the same architecture. The encoder layer is composed of a self-attention layer and a feed-forward neural network while the decoder layer is composed of a self-attention layer, a feed-forward neural network and a cross-attention layer and the other feed-forward neural network.

In each attention layer, following the original Transformer~\cite{transformer}, the input features $X$ are first projected onto these weight matrices to get $Q = XW^Q$, $K = XW^K$ and $V = XW^V$, where the $Q$, $K$, and $V$ denote the triplet of query, key, and value through one $1 \times 1$ convolution on each input features from an image (i.e. either content or style image). Then the output is given by:

\begin{equation}
Attention(Q, K, V) = softmax(\frac{Q  K^T}{\sqrt{d}}) V,
\end{equation}
where the $Q$, $K$, and $V$ denote query, key and
value as described above. All of the three inputs to the attention layer maintain the same dimension $d$.

The detailed network architecture of STTR's transformer could be found in the supplementary.

\subsubsection{Transformer Encoder}

The encoder is used to encode style features, it consists of six encoder layers. Each encoder layer has a standard architecture and consists of a self-attention module and a feed-forward network (FFN). Here, the number of the head is set to be eight. To supplement the image features with position information, fixed positional encodings are added to the flattened features before the features are fed into each attention layer.

\subsubsection{Transformer Decoder}

The decoder is used to reason relationships between content and style tokens while being able to model the global context among them. It consists of two multi-head attention layers and a feed-forward network (FFN). The first multi-head attention layer mainly learns the self-attention within content features while the second one learns the cross-attention between content and style features. 

As shown in Fig.~\ref{fig:framework}, the difference from that of the original Transformer is the input query of the decoder. We adopt content visual tokens as the query. Additionally, we also introduce a fixed positional encoding with the same size of style features.

\subsubsection{CNN Decoder}
The CNN decoder is used to predict the final stylized image. It upsamples the output of the Transformer (with the size of $H_c/8 \times W_c/8 \times 512$) into the original size the same with the input content image. To achieve this, the CNN decoder should upsample it three times. Each upsample module consists of three parts: one standard residual block defined in~\cite{resnet}, one bilinear interpolation layer upsample by a factor of 2, and one $3 \times 3$ convolution. 

Let $RBCk$ denotes a \textbf{R}esidual block-\textbf{B}ilinear interpolation-\textbf{C}onvolution layer where $k$ indicates the channel number of the output features. The CNN decoder architecture consists of $RBC256-RBC128-RBC64-RBC3$.

\vspace{+5pt}
\subsection{Objective Function}\label{sec:loss}
Following existing works~\cite{ulyanov2016texture,ulyanov2017improved,dumoulin2016learned,adain}, we use the pre-trained VGG-19~\cite{simonyan2014very} to compute the loss function to train the STTR.
The loss function $\mathcal{L}$ is defined as a combination of two loss terms, content loss $\mathcal{L}_{c}$ and the style loss $\mathcal{L}_{s}$:
\begin{equation}\label{eqn:loss}
\mathcal{L} = \mathcal{L}_{c} + \lambda \mathcal{L}_{s},
\end{equation}
where $\lambda$ denote the weight of style loss. We empirically set $\lambda$=10.

The content loss $\mathcal{L}_{c}$ is defined by the squared Euclidean distance between the feature representations of the content image $I_c$ and the stylized image $I'$. Thus we define the content loss $\mathcal{L}_{c}$ as follows:

\begin{equation}\label{eqn:content-loss}
\mathcal{L}_{c}=   \left\| f_4({I}_c) -f_4({I'})\right\|_2,
\end{equation}
where we interpret $f_i(\cdot)$ as feature map extracted by a pre-trained VGG-19 network~\cite{simonyan2014very} at the layer $i$. Here, we adopt $relu1\_1$, $relu2\_1$, $relu3\_1$, and $relu4\_1$ as the layers for extracting features.

Similar to~\cite{adain}, the style loss is defined as follows:

\begin{equation}\label{eqn:style-loss}
\begin{split}
\mathcal{L}_{s}&=  \sum_{i=1}^{4} (\left\| \mu(f_i({I}_s)) -\mu(f_i({I'}))\right\|_2 \\
& +\left\| \sigma(f_i({I}_s)) -\sigma(f_i({I'}))\right\|_2),
\end{split}
\end{equation}
where $\mu(\cdot)$, $\sigma(\cdot)$ are the mean and standard deviation, computed across batch size and spatial dimensions independently for each feature channel.

\vspace{+10pt}
\section{Experiments and Discussions}
\subsection{Training Details}
We train our network by using images from MSCOCO~\cite{coco} and WikiArt~\cite{wikiart} as content and style data, respectively. Each dataset contains about 80,000 images. In each training batch, we randomly choose one pair of content and style images with the size of $512 \times 512$ as inputs. We implement our model with PyTorch and apply the Adam optimizer~\cite{adam} with a learning rate of $10^{-5}$.

\begin{figure*}[t]
  \includegraphics[width=1.0\linewidth]{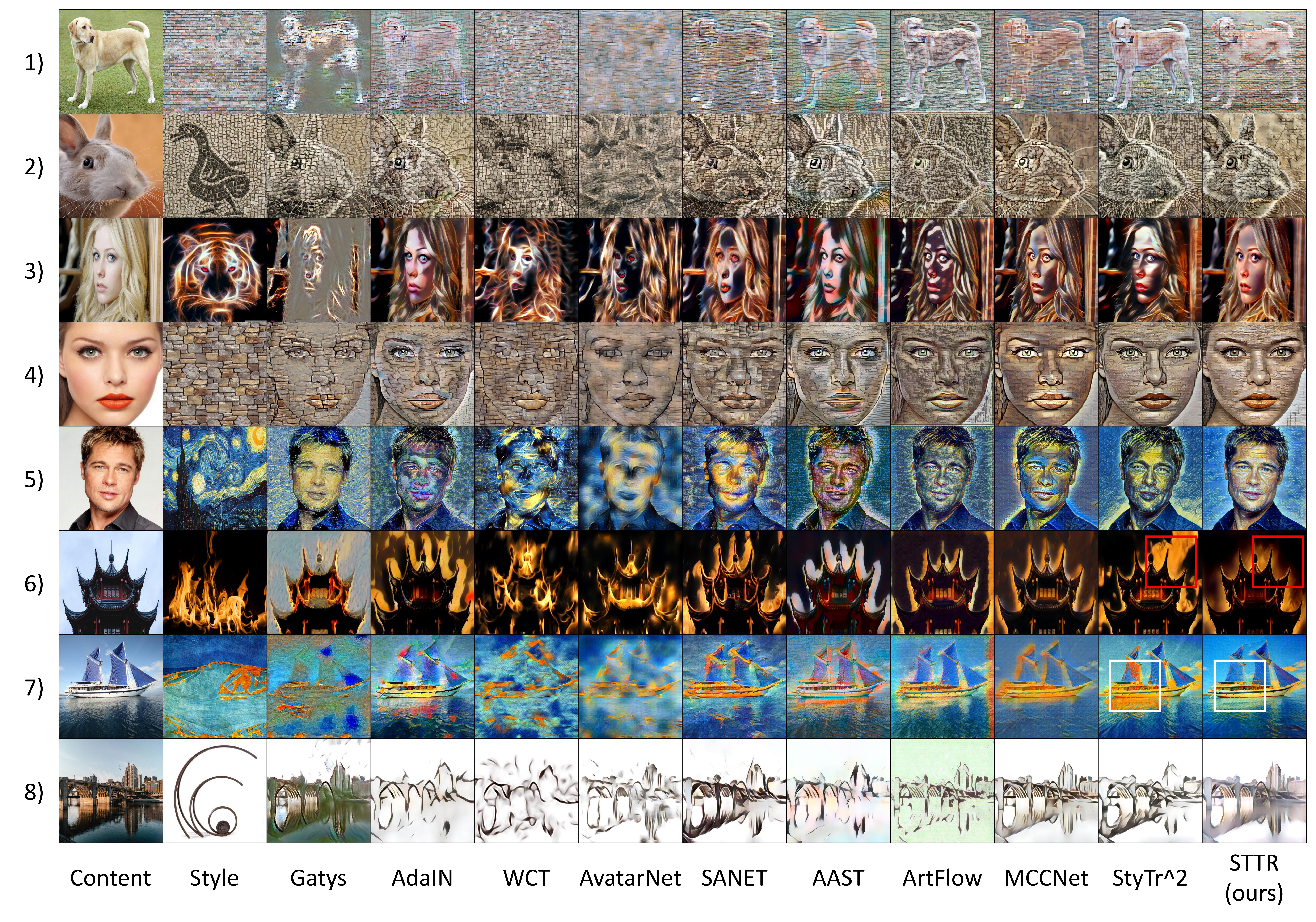}
  \caption{Visual comparison. All content and style images are collected from websites and copyright-free. They are never observed during training. } 
  \label{fig:comparison}
\end{figure*}

\begin{figure}[t]
  \centering
  \includegraphics[width=0.6\linewidth]{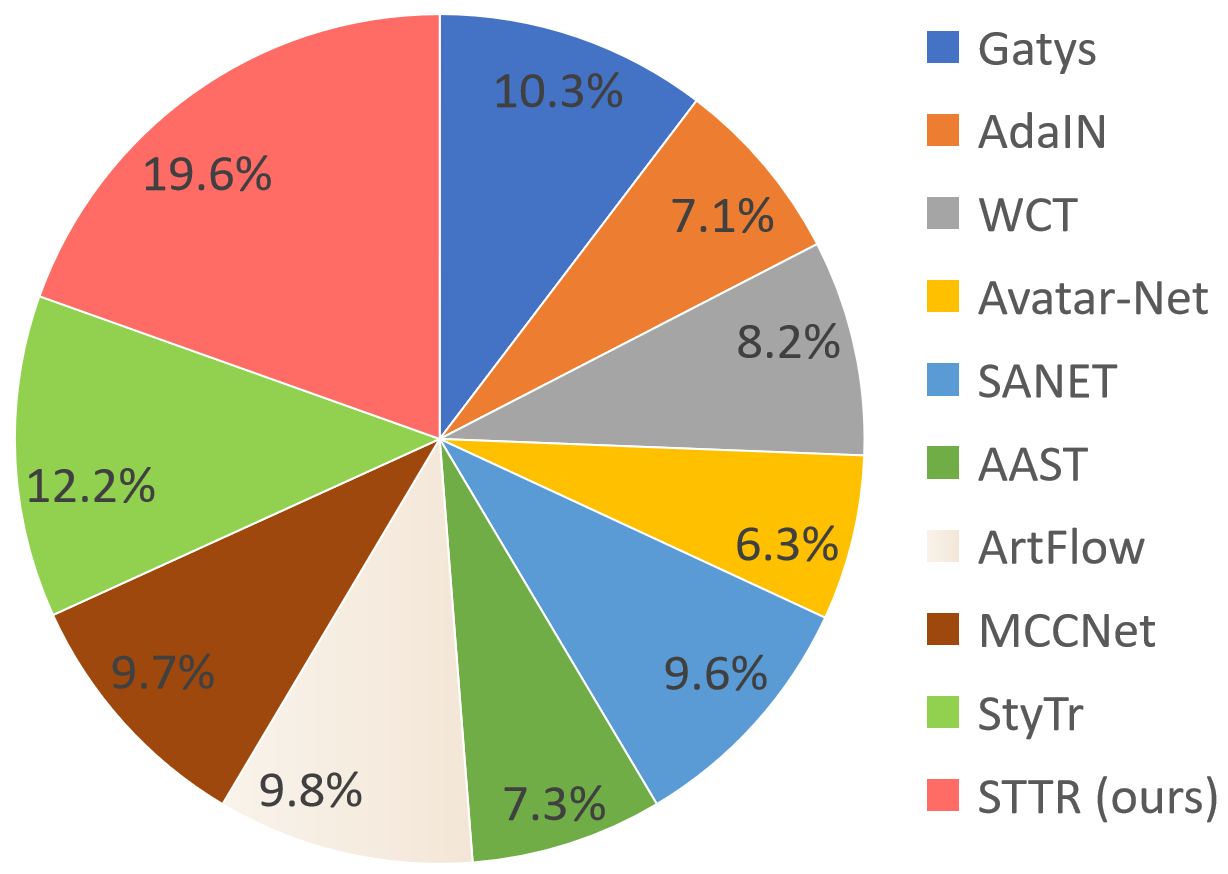}
  \caption{User preference result of seven style transfer algorithms.}
  \label{fig:user-study}
\end{figure}

\vspace{+10pt}
\subsection{Comparison Results}
\subsubsection{Qualitative Evaluations}
There are many quantitative evaluation methods for image quality assessment~\cite{zheng2021learning,wang2004image}. However, as style transfer is a very subjective task and does not have a reference image, to evaluate our method, we first evaluate the quality of the results produced by our model. We visually compare our results to the state-of-the-art style transfer methods: the optimization-based method proposed by Gatys et al.~\cite{gatys2016}, three feature
transformation-based methods WCT~\cite{wct}, AdaIN~\cite{adain}, AAST~\cite{aast}, patch-based method Avatar-Net~\cite{avatar}, attention-based methods (SANET~\cite{sanet} and StyTr~\cite{stytr}), ArtFlow~\cite{artflow}, and MCCNet~\cite{mccnet}. Fig.~\ref{fig:comparison} shows the results. 

As shown in the figure, the proposed STTR performs favorably against the state-of-the-art methods. Both the content structure and style patterns appear well since we match the style tokens onto content tokens in a fine-grained manner.
Although the optimization-based method~\cite{gatys2016} allows universal-style transfer, it is computationally expensive due to its iterative optimization process (see analysis about the speed in Sec.~\ref{sec:efficiency}). Also, the results highly depend on many hyper-parameters, including the number of iterations, the trade-off loss weight, and the initial image. Thus, the results are not robust (e.g., see the $1st$, $3rd$, $6th$, and $7th$ rows in Fig.~\ref{fig:comparison}). The AdaIN~\cite{adain} method presents an efficient solution for arbitrary style transfer, but it generates sub-optimal results (e.g., see the $1st$ and $7th$ rows in Fig.~\ref{fig:comparison}) as it globally stylized the whole content image. The WCT~\cite{wct} and Avatar-Net usually perform well, however, in some cases, they would bring strong distortions which makes it hard to recognize the original object from the style transferred image (e.g., see the $1st$, $2nd$, and $7th$ rows in Fig.~\ref{fig:comparison}). AAST~\cite{aast} transfers color and texture in two independent paths but sometimes cannot produce well-colored results (see the $2nd$ and $4th$ rows in Fig.~\ref{fig:comparison}). In contrast, our method learns semantic relationships between content and style tokens and performs appealing results for arbitrary style, especially working very well on preserving content structure. SANET~\cite{sanet}, ArtFlow~\cite{artflow}, and MCCNet~\cite{mccnet} also hard to generate appealing results especially when the content and style features could not be matched very well (e.g., see the $1st$ and $3rd$ rows in Fig.~\ref{fig:comparison}).

StyTr$^{2}$~\cite{stytr} also adopt a transformer to build a style transfer model and it could produce visually pleasing results. However, it still cannot preserve the fine structures. The main difference is that our method learns correspondence on the semantic-level while StyTr$^{2}$ focuses on the region-level. This is because StyTr$^{2}$ adopts a shallow feature extractor to build the tokenizer. Thus the learned correspondences between the content and style tokens are random. For example, on the $6th$ row, StyTr$^{2}$ generates fires in the sky (marked in the red box) while in our method, the flame area is smoother and appears on the edge of the building. It looks like StyTr$^{2}$ tends to copy-paste the texture in the style image onto the content image. The same thing happens on the $7th$ row. Our model understands the sail and the body of the boat are different objects so the result is clean. However, StyTr$^{2}$ produces results with large yellow areas across the sea and the sail (marked in the white box). It looks like StyTr$^{2}$ tends to copy-paste the texture in the style image onto the content image. Taking advantage of knowledge, our method could learn semantic-level correspondences between the content and style images and reduce distortion.

\vspace{+10pt}
\subsubsection{Perceptual Study} \label{sec:user-study}
As the evaluation of style transfer is highly subjective, we conduct a user study to further evaluate the seven methods shown in Fig.~\ref{fig:comparison}. We use 10 content images and 20 style images. For each method, we use the released codes and default parameters to generate 200 results. 

We hire 50 volunteers on Amazon Mechanical Turk (AMT) for our user study. 
Twenty of 200 content-style pairs are randomly selected for each user. For each style-content pair, we display the stylized results of seven methods on a web page in random order. Each user is asked to vote for the one that he/she likes the most. 

Finally, we collect 1000 votes from 50 users and calculate the percentage of votes that each method received. 

The results are shown in Fig.~\ref{fig:user-study}, where our STTR obtains 19.6\% of the total votes. It is much higher than that of AAST~\cite{aast} whose stylization results are usually thought to be high-quality. Attention-based methods, StyTr$^{2}$~\cite{stytr} and SANET~\cite{sanet} get sub-optimal results. This user study result is consistent with the visual comparisons in Fig.~\ref{fig:comparison} and further demonstrates the superior performance of our proposed STTR.

\begin{table*}
  \caption{Execution time comparison (in seconds).}
  \centering
  \begin{tabular}{cc|c}
    Method & Venue  &Time (512 px)\\
    \hline
    Gatys et al.~\cite{gatys2016} &CVPR16&	106.63\\
AdaIN~\cite{adain}&CVPR17	&0.13\\
WCT~\cite{wct}&CVPR17	&1.63\\
Avatar-Net~\cite{avatar}&CVPR18&	0.49\\
SANET~\cite{sanet}&CVPR19	&0.19\\
AAST~\cite{aast}&ACMMM20	&0.49\\
ArtFlow~\cite{artflow}&CVPR21	&0.40\\
MCCNet~\cite{mccnet}&AAAI21	&0.15\\
StyTr$^2$~\cite{stytr}&CVPR22	&0.57\\
    \hline
STTR (ours) &0.14&	0.28\\
  \end{tabular}
  \label{tab:time}
\end{table*}

\subsubsection{Efficiency}\label{sec:efficiency}
Table~\ref{tab:time} shows the run time performance of the proposed method and other methods at 512 pixels image scales. All the methods are tested on a PC with
an Intel(R) Xeon(R) Gold 6226R 2.90GHz 
CPU 
and a Tesla V100 
GPU.
The optimization-based method~\cite{gatys2016} is impractically computationally expensive because of its iterative optimization process. 
WCT~\cite{wct} is also relatively slower as it conducts multi-level coarse-to-fine stylization.

In contrast, our proposed STTR runs at 4 frames per second (fps) for $512 \times 512$ images. Therefore, our method could feasibly process style transfer in real time. Our model is much faster than Gatys et al.~\cite{gatys2016} and achieves better qualitative performance as shown in Fig.~\ref{fig:comparison}.

\subsubsection{Memory usage}
Compared with two current SOTAs, our method has fewer parameters than StyTr$^{2}$ (ours: 45.643M v.s. StyTr$^{2}$: 48.339M) and takes less memory than AdaAttn (ours: 18391M v.s. AdaAttn: 25976M) on a $512 \times 512$ image. However, such memory costs could be further reduced by using fewer channel numbers.

\subsubsection{Results for Video Style Transfer}\label{sec:video-res}
\begin{figure*}[t]
  \includegraphics[width=1.0\linewidth]{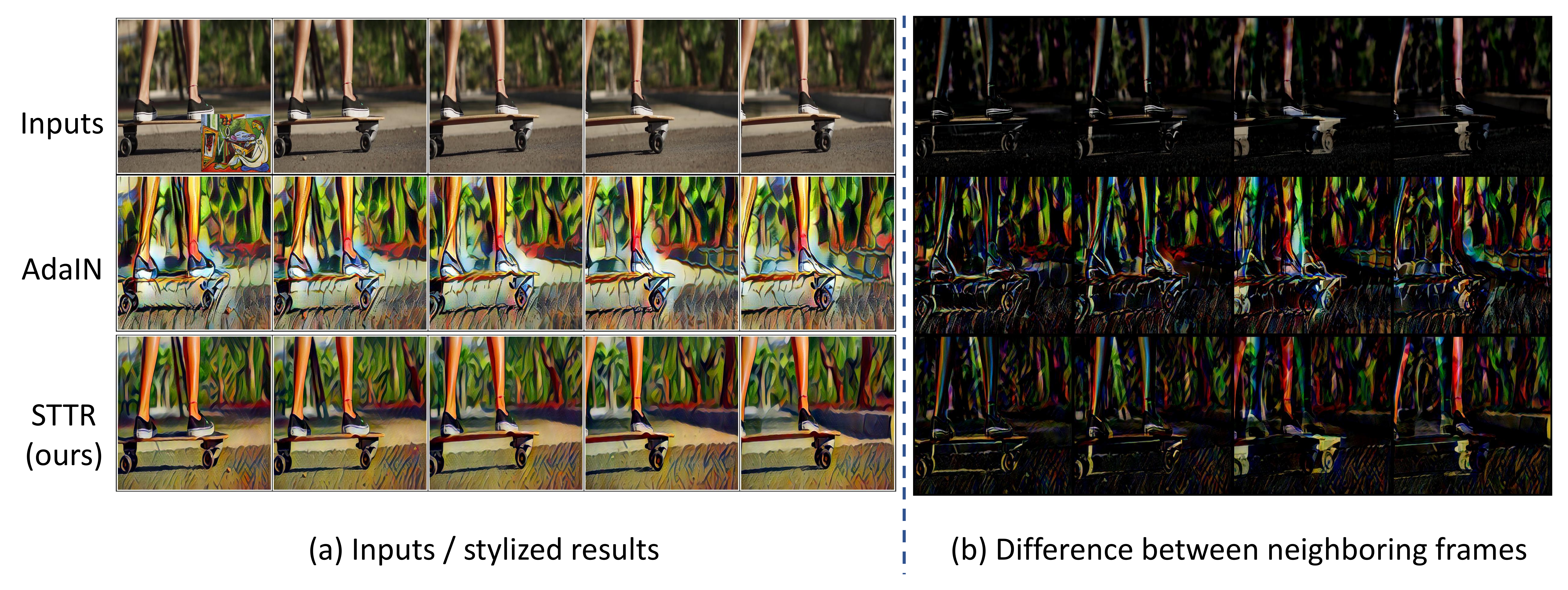}
  \caption{Visual comparison for video style transfer. }
  
  \label{fig:supp-comparison-video}
\end{figure*}

Here we also provide results of video style transfer. As shown in Fig.~\ref{fig:supp-comparison-video}, our model can perform video stylization on video sequence frame-by-frame. On the contrary, AdaIN~\cite{adain} produces results with high motion blur. To show the stable results produced by our model, we further compute the difference between neighboring frames to show the smoothness between frames. It is clearly shown that the difference generated by our method is much more close to that from the input frames. It is because our method could well preserve the content structure and reduce the motion-blurred video frames.

\subsection{Ablations}
To verify the effect of each component in the STTR, we conduct an ablation study by visually comparing the generated results without each of them. All of the compared models are trained by the same 400,000 content-style pairs (i.e., trained for 5 epochs).
More ablation experiments could be found in the supplementary.

\begin{figure}[t]
  \includegraphics[width=1.0\linewidth]{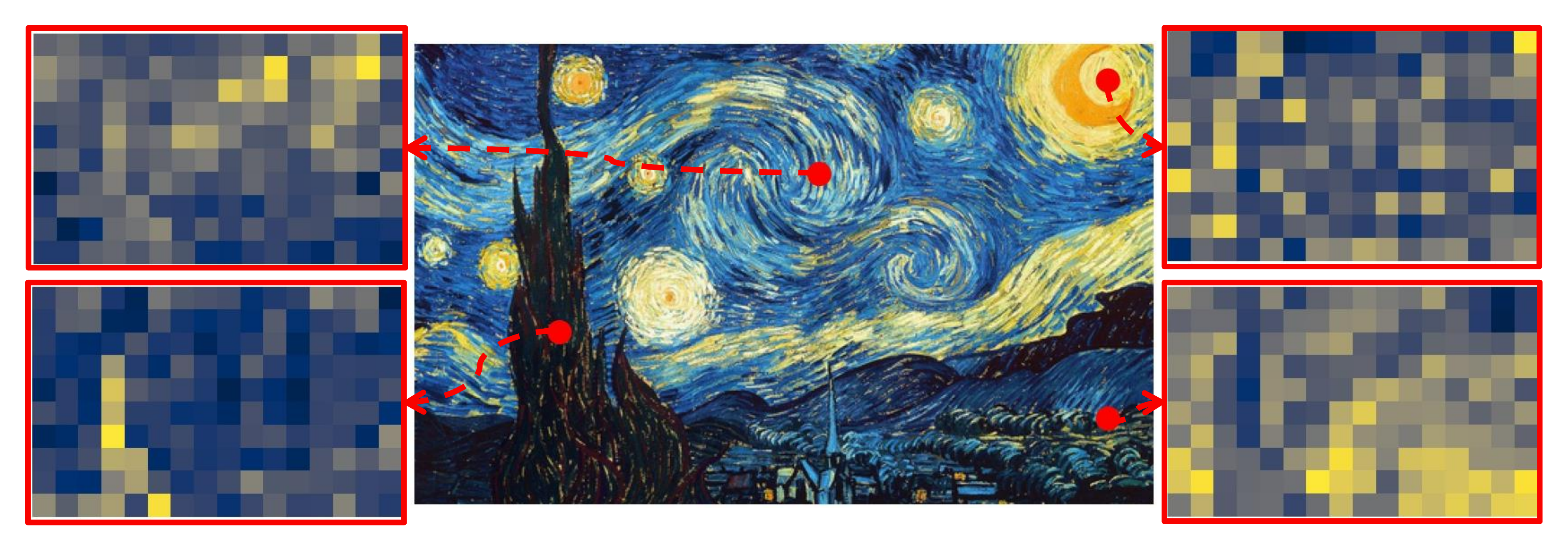}
  \caption{Encoder attention maps for different sampling points on style image. The style image (e.g., Van Gogh's painting) is shown in the center. We further show four sampled points and their according attention maps with different colors. 
  }
  
  \label{fig:enc-attention}

\end{figure}

\begin{figure}[t]
  \includegraphics[width=1.0\linewidth]{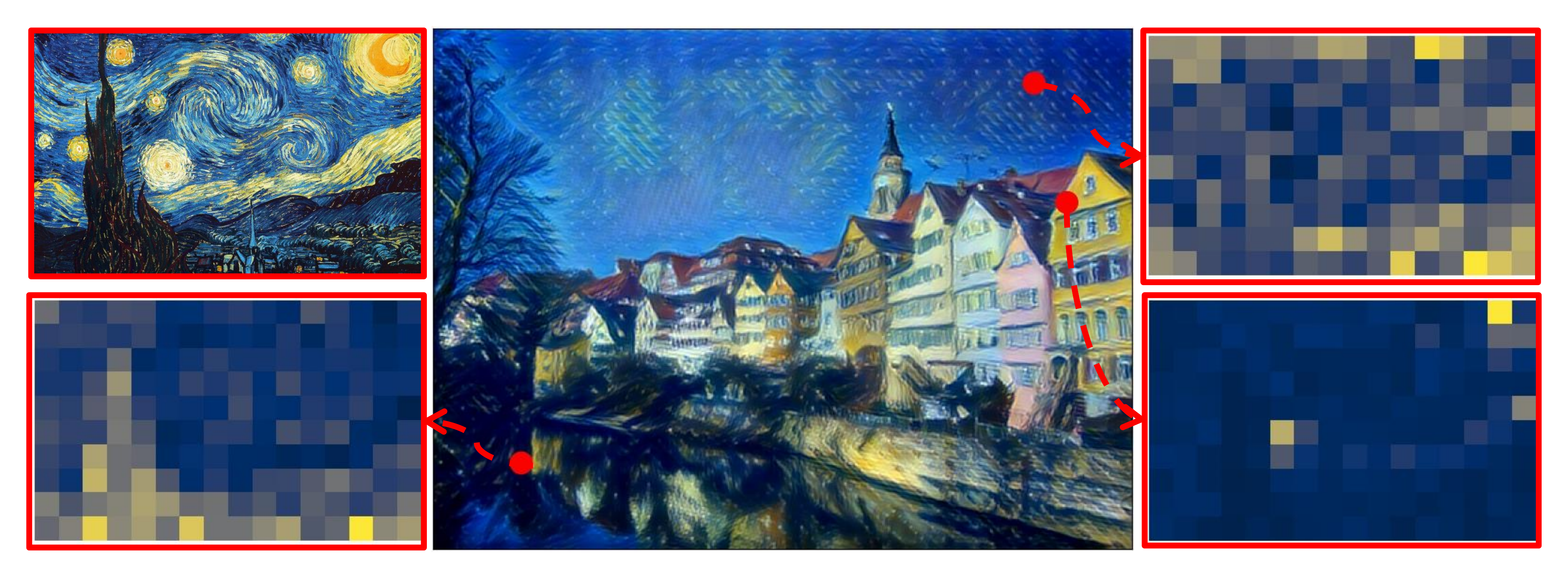}
  \caption{Decoder attention maps for different sampling points on the output. The stylized image is shown in the center. We further show three sampled points and their according attention maps with different colors. On the upper left is the input style image.
  }
  \label{fig:dec-attention}
\end{figure}

\subsubsection{Learned Attention Maps}\label{sec:ablation-encoder}

We can observe model with deeper encoder and decoder has stronger capability to preserve semantic similarity, so that similar style patterns (e.g., fire) can be transferred to similar content regions. 

The encoder is used for encoding the style pattern while the decoder learns a matching between the content and style features.

In Fig.~\ref{fig:enc-attention}, we provide visualized attention maps to show that self-attention could encode style features and similar visual tokens can be grouped together. For the top right marked point (which locates the center of the sun), all of the regions related to other suns have a higher value (in yellow color) in their corresponding attention map. In Fig.~\ref{fig:dec-attention}, attention maps show the learned fine-grained relationships between content and style features. The lower left point in the content image has strong relationships (e.g., see its attention map) with the black tree in the style image. So the pixels around that point in the output are very dark (see the stylized results located around that point).

\subsubsection{The Loss Weight $\lambda$} \label{sec:ablation-loss-weight}
\begin{figure}[t]
  \centering
  \includegraphics[width=0.9\linewidth]{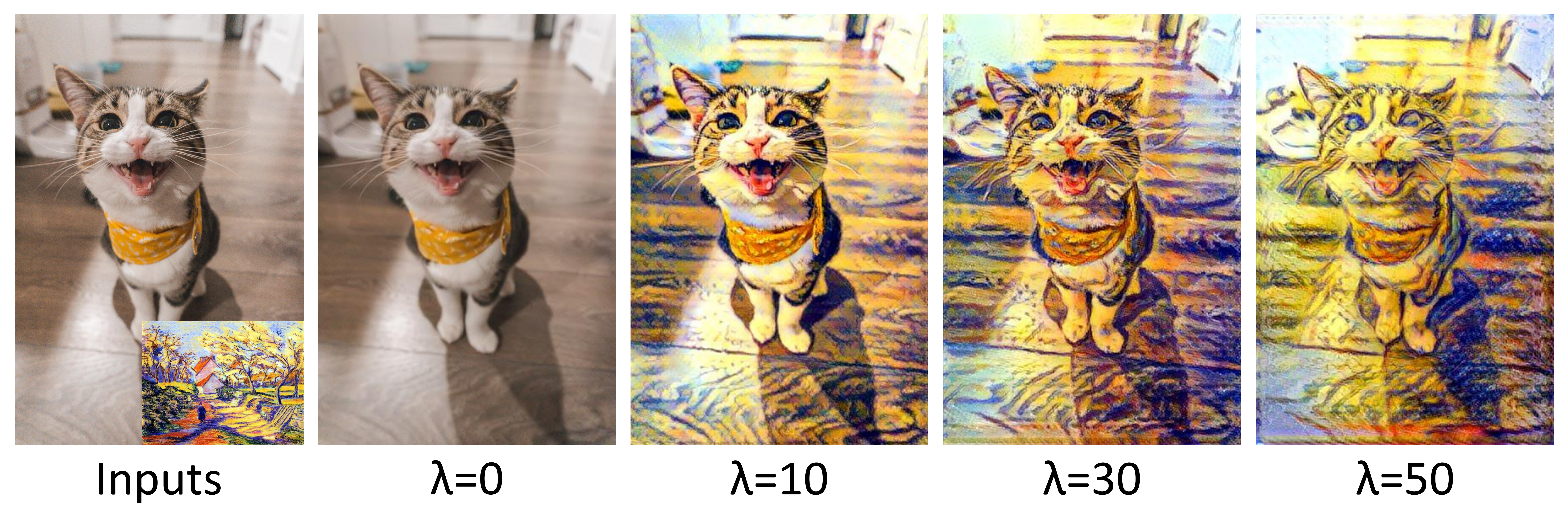}
  \caption{Trading off between the content and style images. 
  \label{fig:loss-weight}
}
\end{figure}

The degree of stylization can be modified by the hyper-parameter $\lambda$ as described in Sec.~\ref{sec:loss}. As shown in Fig.~\ref{fig:loss-weight}, changing $\lambda$ in the training stage could control the magnitude of stylization.

\vspace{+10pt}
\section{Conclusion}
Image style transfer mainly aims to transfer style patterns into the content images. Taking advantage of the breakthrough of vision Transformers, in this work, we propose a style Transformer for solving image style transfer tasks. Since STTR breaks content and style features into tokens and computes very local relationships between them, fine-grained style transformation could be obtained. Experiments demonstrate that the proposed model can produce visually pleasing results and preserve content structure very well without bringing a heavy timing cost. The proposed method also has the potential to be applied to video style transfer frameworks. There are still some limitations that call for continuing efforts. In the future, we would like to extend STTR to handle multiple style mixtures and work on light-weight Transformer architecture for style transfer.

\newpage
\bibliographystyle{splncs}
\bibliography{reference}

\begin{thebibliography}{10}

\bibitem{clothes}
Date, P., Ganesan, A., Oates, T.:
\newblock Fashioning with networks: neural style transfer to design clothes.
\newblock In: KDD ML4Fashion workshop. Volume~2. (2017)

\bibitem{video}
Chen, D., Liao, J., Yuan, L., Yu, N., Hua, G.:
\newblock Coherent online video style transfer.
\newblock In: ICCV. (2017)  1105--1114

\bibitem{zhang2013style}
Zhang, W., Cao, C., Chen, S., Liu, J., Tang, X.:
\newblock Style transfer via image component analysis.
\newblock TMM \textbf{15} (2013)  1594--1601

\bibitem{liu2017photo}
Liu, J., Yang, W., Sun, X., Zeng, W.:
\newblock {Photo Stylistic Brush}: Robust style transfer via superpixel-based
  bipartite graph.
\newblock TMM \textbf{20} (2017)  1724--1737

\bibitem{virtusio2021neural}
Virtusio, J.J., Ople, J.J.M., Tan, D.S., Tanveer, M., Kumar, N., Hua, K.L.:
\newblock {Neural Style Palette: A} multimodal and interactive style transfer
  from a single style image.
\newblock TMM (2021)

\bibitem{material}
Matsuo, S., Shimoda, W., Yanai, K.:
\newblock Partial style transfer using weakly supervised semantic segmentation.
\newblock In: ICME Workshops, IEEE (2017)  267--272

\bibitem{kim2019style}
Kim, B.K., Kim, G., Lee, S.Y.:
\newblock {Style-Controlled} synthesis of clothing segments for fashion image
  manipulation.
\newblock TMM \textbf{22} (2019)  298--310

\bibitem{liu2019swapgan}
Liu, Y., Chen, W., Liu, L., Lew, M.S.:
\newblock Swapgan: A multistage generative approach for person-to-person
  fashion style transfer.
\newblock TMM \textbf{21} (2019)  2209--2222

\bibitem{vr}
Castillo, C., De, S., Han, X., Singh, B., Yadav, A.K., Goldstein, T.:
\newblock {Son of Zorn's Lemma: Targeted} style transfer using instance-aware
  semantic segmentation.
\newblock In: ICASSP, IEEE (2017)  1348--1352

\bibitem{gatys2016}
Gatys, L.A., Ecker, A.S., Bethge, M.:
\newblock Image style transfer using convolutional neural networks.
\newblock In: CVPR. (2016)  2414--2423

\bibitem{johnson2016perceptual}
Johnson, J., Alahi, A., Fei-Fei, L.:
\newblock Perceptual losses for real-time style transfer and super-resolution.
\newblock In: ECCV, Springer (2016)  694--711

\bibitem{li2016precomputed}
Li, C., Wand, M.:
\newblock Precomputed real-time texture synthesis with markovian generative
  adversarial networks.
\newblock In: ECCV, Springer (2016)  702--716

\bibitem{ulyanov2016texture}
Ulyanov, D., Lebedev, V., Vedaldi, A., Lempitsky, V.S.:
\newblock {Texture Networks: Feed-forward} synthesis of textures and stylized
  images.
\newblock In: ICML. (2016) ~4

\bibitem{dumoulin2016learned}
Dumoulin, V., Shlens, J., Kudlur, M.:
\newblock A learned representation for artistic style.
\newblock arXiv preprint arXiv:1610.07629 (2016)

\bibitem{chen2017stylebank}
Chen, D., Yuan, L., Liao, J., Yu, N., Hua, G.:
\newblock Stylebank: An explicit representation for neural image style
  transfer.
\newblock In: CVPR. (2017)  1897--1906

\bibitem{li2017diversified}
Li, Y., Fang, C., Yang, J., Wang, Z., Lu, X., Yang, M.H.:
\newblock Diversified texture synthesis with feed-forward networks.
\newblock In: CVPR. (2017)  3920--3928

\bibitem{zhang2018multi}
Zhang, H., Dana, K.:
\newblock Multi-style generative network for real-time transfer.
\newblock In: ECCV Workshops. (2018)  0--0

\bibitem{adain}
Huang, X., Belongie, S.:
\newblock Arbitrary style transfer in real-time with adaptive instance
  normalization.
\newblock In: ICCV. (2017)  1501--1510

\bibitem{wct}
Li, Y., Fang, C., Yang, J., Wang, Z., Lu, X., Yang, M.H.:
\newblock Universal style transfer via feature transforms.
\newblock arXiv preprint arXiv:1705.08086 (2017)

\bibitem{kitov2019depth}
Kitov, V., Kozlovtsev, K., Mishustina, M.:
\newblock {Depth-Aware Arbitrary} style transfer using instance normalization.
\newblock arXiv preprint arXiv:1906.01123 (2019)

\bibitem{aast}
Hu, Z., Jia, J., Liu, B., Bu, Y., Fu, J.:
\newblock {Aesthetic-Aware} image style transfer.
\newblock In: ACM MM. (2020)  3320--3329

\bibitem{stytr}
Deng, Y., Tang, F., Pan, X., Dong, W., Ma, C., Xu, C.:
\newblock Stytr\^{}2: Unbiased image style transfer with transformers.
\newblock CVPR (2021)

\bibitem{transformer}
Vaswani, A., Shazeer, N., Parmar, N., Uszkoreit, J., Jones, L., Gomez, A.N.,
  Kaiser, L., Polosukhin, I.:
\newblock Attention is all you need.
\newblock arXiv preprint arXiv:1706.03762 (2017)

\bibitem{mahendran2015understanding}
Mahendran, A., Vedaldi, A.:
\newblock Understanding deep image representations by inverting them.
\newblock In: CVPR. (2015)  5188--5196

\bibitem{gatys2015neural}
Gatys, L.A., Ecker, A.S., Bethge, M.:
\newblock A neural algorithm of artistic style.
\newblock arXiv preprint arXiv:1508.06576 (2015)

\bibitem{chen2016fast}
Chen, T.Q., Schmidt, M.:
\newblock Fast patch-based style transfer of arbitrary style.
\newblock arXiv preprint arXiv:1612.04337 (2016)

\bibitem{artflow}
An, J., Huang, S., Song, Y., Dou, D., Liu, W., Luo, J.:
\newblock Artflow: Unbiased image style transfer via reversible neural flows.
\newblock In: CVPR. (2021)  862--871

\bibitem{mccnet}
Deng, Y., Tang, F., Dong, W., Huang, H., Ma, C., Xu, C.:
\newblock Arbitrary video style transfer via multi-channel correlation.
\newblock In: AAAI. (2021)  1210--1217

\bibitem{sanet}
Park, D.Y., Lee, K.H.:
\newblock Arbitrary style transfer with style-attentional networks.
\newblock In: CVPR. (2019)  5880--5888

\bibitem{adaattn}
Liu, S., Lin, T., He, D., Li, F., Wang, M., Li, X., Sun, Z., Li, Q., Ding, E.:
\newblock {AdaAttN: Revisit} attention mechanism in arbitrary neural style
  transfer.
\newblock In: ICCV. (2021)  6649--6658

\bibitem{hong2021domain}
Hong, K., Jeon, S., Yang, H., Fu, J., Byun, H.:
\newblock {Domain-Aware Universal Style Transfer}.
\newblock In: ICCV. (2021)

\bibitem{vit}
Dosovitskiy, A., Beyer, L., Kolesnikov, A., Weissenborn, D., Zhai, X.,
  Unterthiner, T., Dehghani, M., Minderer, M., Heigold, G., Gelly, S.,  et~al.:
\newblock {An Image is Worth 16x16 Words: Transformers} for image recognition
  at scale.
\newblock arXiv preprint arXiv:2010.11929 (2020)

\bibitem{detr}
Carion, N., Massa, F., Synnaeve, G., Usunier, N., Kirillov, A., Zagoruyko, S.:
\newblock {End-to-End Object Detection with Transformers}.
\newblock In: ECCV, Springer (2020)  213--229

\bibitem{beal2020toward}
Beal, J., Kim, E., Tzeng, E., Park, D.H., Zhai, A., Kislyuk, D.:
\newblock Toward transformer-based object detection.
\newblock arXiv preprint arXiv:2012.09958 (2020)

\bibitem{pan20203d}
Pan, X., Xia, Z., Song, S., Li, L.E., Huang, G.:
\newblock 3d object detection with pointformer.
\newblock arXiv preprint arXiv:2012.11409 (2020)

\bibitem{yuan2020temporal}
Yuan, Z., Song, X., Bai, L., Zhou, W., Wang, Z., Ouyang, W.:
\newblock {Temporal-Channel Transformer for 3D Lidar-Based Video Object
  Detection in Autonomous Driving}.
\newblock arXiv preprint arXiv:2011.13628 (2020)

\bibitem{zhu2020deformable}
Zhu, X., Su, W., Lu, L., Li, B., Wang, X., Dai, J.:
\newblock {Deformable DETR: Deformable Transformers for End-to-End Object
  Detection}.
\newblock arXiv preprint arXiv:2010.04159 (2020)

\bibitem{wang2020max}
Wang, H., Zhu, Y., Adam, H., Yuille, A., Chen, L.C.:
\newblock {MaX-DeepLab: End-to-End Panoptic Segmentation with Mask
  Transformers}.
\newblock arXiv preprint arXiv:2012.00759 (2020)

\bibitem{wang2020end}
Wang, Y., Xu, Z., Wang, X., Shen, C., Cheng, B., Shen, H., Xia, H.:
\newblock {End-to-End Video Instance Segmentation with Transformers}.
\newblock arXiv preprint arXiv:2011.14503 (2020)

\bibitem{zheng2020rethinking}
Zheng, S., Lu, J., Zhao, H., Zhu, X., Luo, Z., Wang, Y., Fu, Y., Feng, J.,
  Xiang, T., Torr, P.H.,  et~al.:
\newblock Rethinking semantic segmentation from a sequence-to-sequence
  perspective with transformers.
\newblock arXiv preprint arXiv:2012.15840 (2020)

\bibitem{huang2020hand}
Huang, L., Tan, J., Liu, J., Yuan, J.:
\newblock {Hand-Transformer}: Non-autoregressive structured modeling for 3d
  hand pose estimation.
\newblock In: ECCV, Springer (2020)  17--33

\bibitem{huang2020hot}
Huang, L., Tan, J., Meng, J., Liu, J., Yuan, J.:
\newblock {HOT-Net: Non-Autoregressive Transformer for 3D Hand-Object Pose
  Estimation}.
\newblock In: ACM MM. (2020)  3136--3145

\bibitem{lin2020end}
Lin, K., Wang, L., Liu, Z.:
\newblock {End-to-End Human Pose and Mesh Reconstruction with Transformer}.
\newblock arXiv preprint arXiv:2012.09760 (2020)

\bibitem{yang2020transpose}
Yang, S., Quan, Z., Nie, M., Yang, W.:
\newblock {TransPose: Towards Explainable Human Pose Estimation by
  Transformer}.
\newblock arXiv preprint arXiv:2012.14214 (2020)

\bibitem{igpt}
Chen, M., Radford, A., Child, R., Wu, J., Jun, H., Luan, D., Sutskever, I.:
\newblock Generative pretraining from pixels.
\newblock In: ICML, PMLR (2020)  1691--1703

\bibitem{gptv2}
Radford, A., Wu, J., Child, R., Luan, D., Amodei, D., Sutskever, I.:
\newblock Language models are unsupervised multitask learners.
\newblock OpenAI blog \textbf{1} (2019) ~9

\bibitem{zeng2021improving}
Zeng, Y., Yang, H., Chao, H., Wang, J., Fu, J.:
\newblock Improving visual quality of image synthesis by a token-based
  generator with transformers.
\newblock In: NeurIPS. (2021)

\bibitem{ttsr}
Yang, F., Yang, H., Fu, J., Lu, H., Guo, B.:
\newblock Learning texture transformer network for image super-resolution.
\newblock In: CVPR. (2020)  5791--5800

\bibitem{liu2022learning}
Liu, C., Yang, H., Fu, J., Qian, X.:
\newblock Learning trajectory-aware transformer for video super-resolution.
\newblock In: CVPR. (2022)

\bibitem{qiu2022learning}
Qiu, Z., Yang, H., Fu, J., Fu, D.:
\newblock Learning spatiotemporal frequency-transformer for compressed video
  super-resolution.
\newblock arXiv preprint arXiv:2208.03012 (2022)

\bibitem{liu2022ttvfi}
Liu, C., Yang, H., Fu, J., Qian, X.:
\newblock {TTVFI: Learning Trajectory-Aware Transformer for Video Frame
  Interpolation}.
\newblock arXiv preprint arXiv:2207.09048 (2022)

\bibitem{resnet}
He, K., Zhang, X., Ren, S., Sun, J.:
\newblock Deep residual learning for image recognition.
\newblock In: CVPR. (2016)  770--778

\bibitem{ulyanov2017improved}
Ulyanov, D., Vedaldi, A., Lempitsky, V.:
\newblock {Improved Texture Networks: Maximizing Quality and Diversity in
  Feed-forward Stylization and Texture Synthesis}.
\newblock In: CVPR. (2017)  6924--6932

\bibitem{simonyan2014very}
Simonyan, K., Zisserman, A.:
\newblock Very deep convolutional networks for large-scale image recognition.
\newblock arXiv preprint arXiv:1409.1556 (2014)

\bibitem{coco}
Lin, T.Y., Maire, M., Belongie, S., Hays, J., Perona, P., Ramanan, D.,
  Doll{\'a}r, P., Zitnick, C.L.:
\newblock {Microsoft COCO: Common Objects in Context}.
\newblock In: ECCV, Springer (2014)  740--755

\bibitem{wikiart}
Nichol, K.:
\newblock Painter by numbers, wikiart (2016)

\bibitem{adam}
Kingma, D.P., Ba, J.:
\newblock {Adam: A Method for Stochastic Optimization}.
\newblock arXiv preprint arXiv:1412.6980 (2014)

\bibitem{zheng2021learning}
Zheng, H., Yang, H., Fu, J., Zha, Z.J., Luo, J.:
\newblock Learning conditional knowledge distillation for degraded-reference
  image quality assessment.
\newblock In: ICCV. (2021)

\bibitem{wang2004image}
Wang, Z., Bovik, A.C., Sheikh, H.R., Simoncelli, E.P.:
\newblock Image quality assessment: from error visibility to structural
  similarity.
\newblock TIP \textbf{13} (2004)  600--612

\bibitem{avatar}
Sheng, L., Lin, Z., Shao, J., Wang, X.:
\newblock {Avatar-Net: Multi-scale Zero-shot Style Transfer by Feature
  Decoration}.
\newblock In: CVPR. (2018)  8242--8250

\bibitem{stroke}
Jing, Y., Liu, Y., Yang, Y., Feng, Z., Yu, Y., Tao, D., Song, M.:
\newblock {Stroke Controllable Fast Style Transfer with Adaptive Receptive
  Fields}.
\newblock In: ECCV. (2018)  238--254

\end{thebibliography}
\begin{appendix}

\title{Fine-Grained Image Style Transfer with Visual Transformers - Supplementary Material}

\author{Jianbo~Wang\inst{1} \and
        Huan~Yang\inst{2} \and
        Jianlong~Fu\inst{2} \and \\
        Toshihiko~Yamasaki\inst{1} \and
        Baining~Guo\inst{2}
}
\authorrunning{J. Wang et al.}
%
\institute{The Univerisity of Tokyo \email{\{jianbowang815,yamasaki\}@cvm.t.u-tokyo.ac.jp}
\and Microsoft Research \email{\{huayan,jianf,bainguo\}@microsoft.com}
}
\maketitle              

\section{Architecture of STTR's Transformer}
\subsection{Tokenizer}
In our model, one image is divided into a set of visual tokens. Thus, we have to first convert the input image into a set of visual tokens. We assume that each of them represents a semantic concept in the image. We then feed these tokens to a transformer.
Let us denote the input feature map by $X\in \mathbb{R}^{H \times W \times C}$ (height $H$, width $W$, and channels $C$) and visual tokens by $T\in \mathbb{R}^{L \times C}$ where $L$ indicates the number of tokens.

In the main paper, we adopt a filter-based tokenizer. Here we would like to compare the two tokenizers (i.e., unfold-based tokenizer and filter-based tokenizer) and explain why we choose the filter-based tokenizer.

\begin{figure}[t]
  \includegraphics[width=1.0\linewidth]{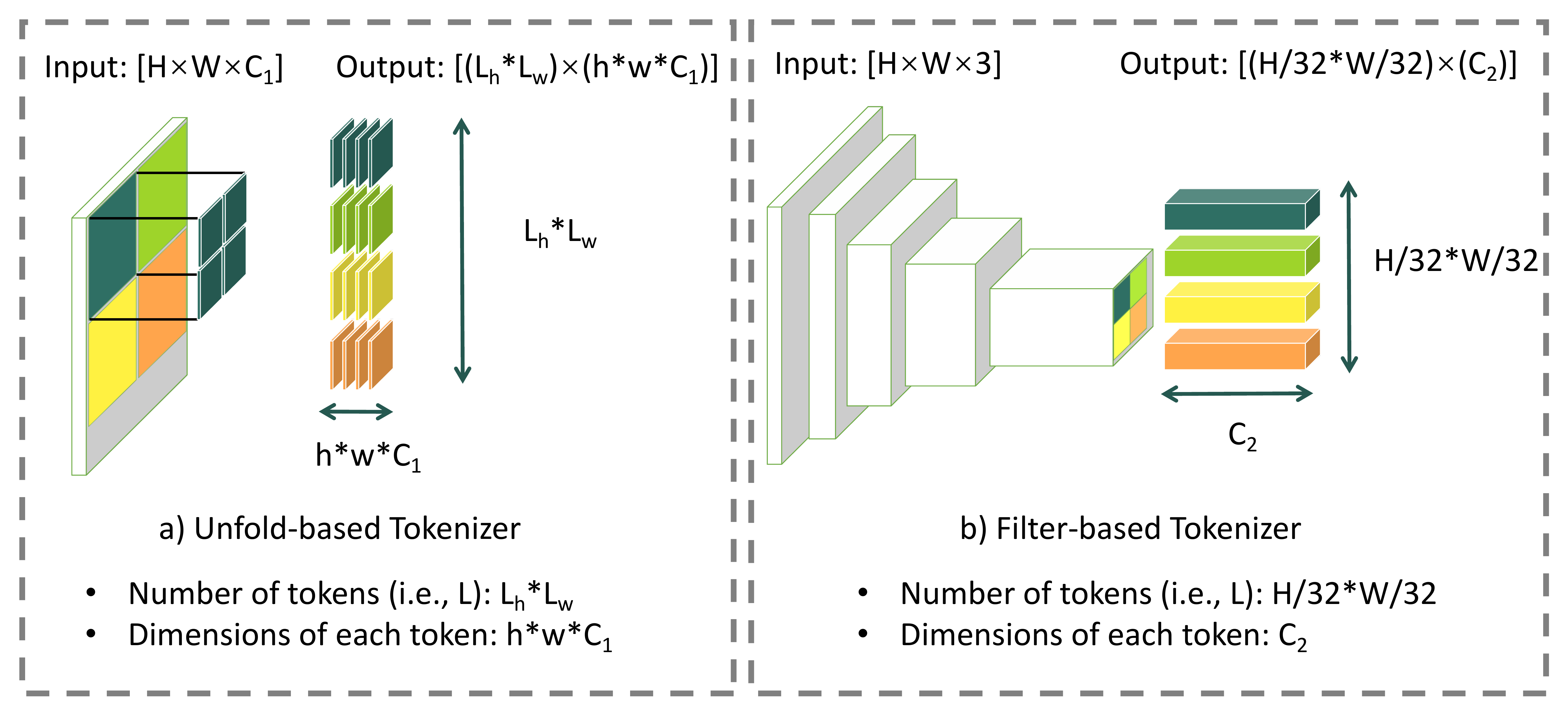}
  \caption{Illustration of different dimension configurations of the backbone. (a)~Unfold-based tokenizer maintains spatial dimensions of features and then unfolds them into patches; (b)~Filter-based tokenizer gradually downsamples the features using ResNet-50 to obtain spatially smaller output features.  }
  
  \label{fig:spa-dim}
\end{figure}

\textbf{Unfold-based Tokenizer.}
The first choice is to unfold the features by using a sliding window to slide along $H$ and $W$ dimensions to create blocks (always with overlaps). Let us assume the output size of extracted feature maps is $H \times W \times C_1$, where $H$, $W$, and $C_1$ indicate height, width, and the number of channels. As shown in Fig.~\ref{fig:spa-dim}(a), we directly slide local blocks from image features into patches. If the shape of each patch is $h \times w$ (height of each patch $h$, width of each patch $w$), then we could obtain $L$ tokens.  For each token, the dimension is $h \times w \times C_1$. $L$ could be calculated as follows:

\begin{equation}\label{eqn:cal_l}
L = \prod_d \left\lfloor\frac{spatial\_size[d]-kernel\_size[d]}{stride[d]}+1\right\rfloor,
\end{equation}
where $d$ is overall spatial dimensions, $spatial\_size$, $kernel\_size$, and $stride$ are formed by the spatial dimensions of input, the size of the convolution kernels, and the stride for the sliding window. $\lfloor x \rfloor$ function calculates the largest integer that is less than or equal to $x$. Usually, $L \ll H \times W$. Specific to our model, such a design may result in strong blocky artifacts. This is because we reshape the feature from the spatial dimension into the channel dimension and the style loss will constrain the predicted results to match the mean and variance with the target style features in each channel.

\subsubsection{Transformer Encoder}
\begin{figure}[t]
  \centering
  \includegraphics[width=0.7\linewidth]{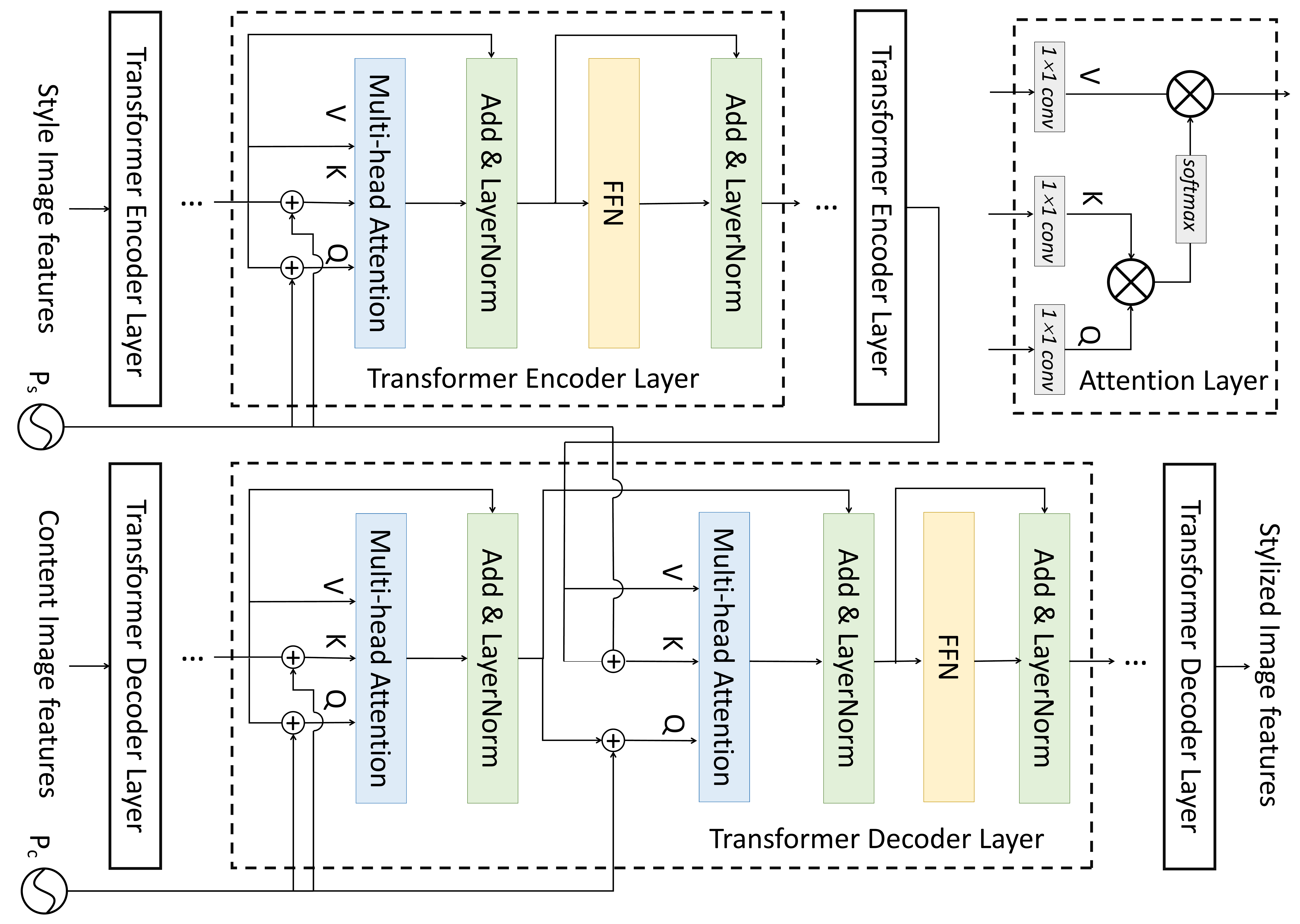}
  \caption{Architecture of STTR's Transformer, including Transformer encoder and decoder. The $\oplus$  and $\otimes$ represent matrix addition and dot-product operations, respectively. $P_s$ and $P_c$ represent the positional encoding of style and content features, respectively.
}
  \label{fig:trans-enc-dec}
\end{figure}

The detailed architecture of STTR's Transformer could be found in Fig.~\ref{fig:trans-enc-dec}.

\begin{figure}[t]
  \includegraphics[width=1.0\linewidth]{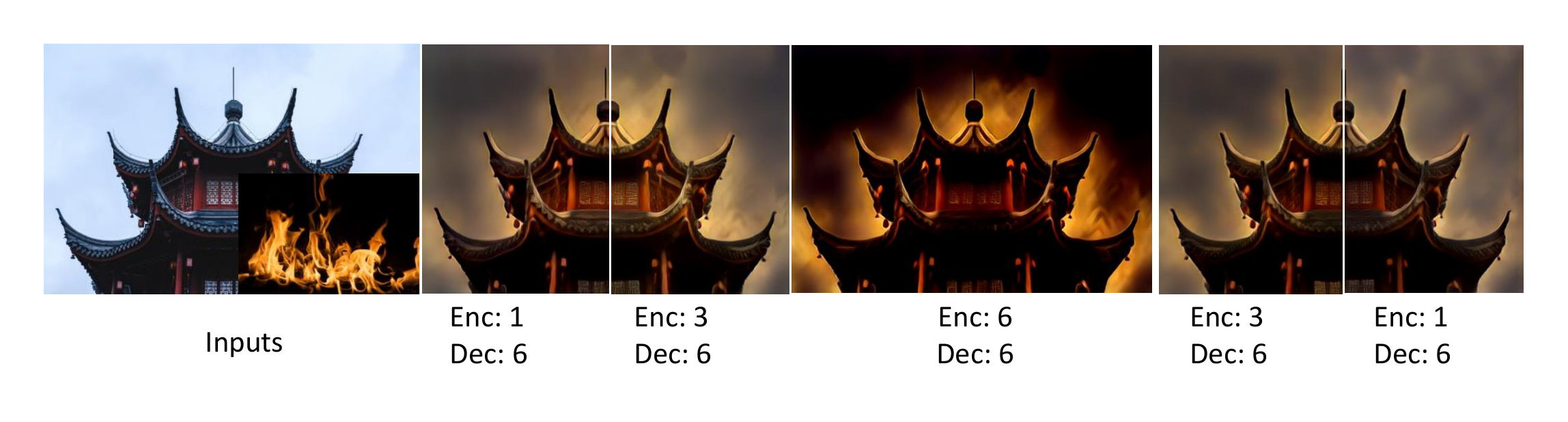}
  \caption{Effects of different encoder and decoder sizes. ``Eec'' and ``Dec'' indicate the number of encoder and decoder layer respectively.
  }
  
  \label{fig:enc-dec-number}
\end{figure}

\section{Ablations}

\subsubsection{The Tokenizer} 
\begin{figure}[t]
  \includegraphics[width=1.0\linewidth]{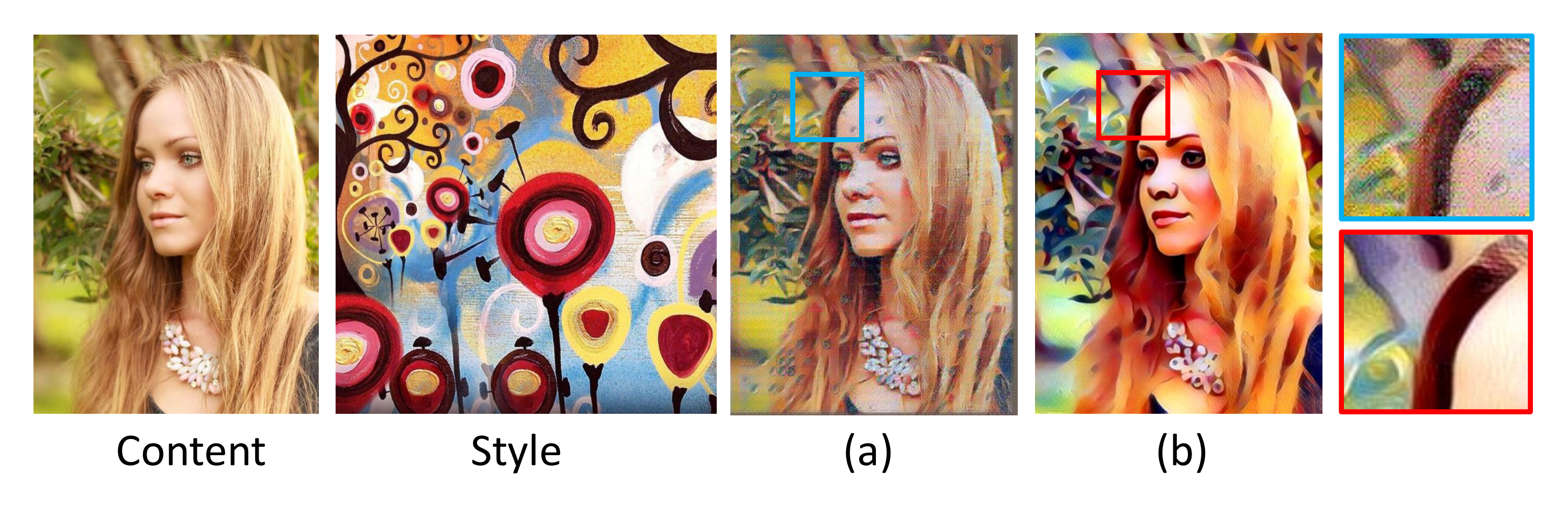}
  \caption{Visualization of outputs from different dimension configurations of the backbone. (a)~Unfold-based tokenizer; (b)~Filter-based tokenizer. Zoom in for a better view.}
  \label{fig:spa-vis}
\end{figure}

We compare the outputs from different dimension configurations of the backbone. As shown in Fig.~\ref{fig:spa-vis}(a), the result from the unfold-based tokenizer has strong blocky artifacts. This is because the model only sees patches and could thus create artifacts on the borders of these patches. Also, the supervision signal from the style loss will guide the predicted results to match the mean and variance with the target style features in each channel. This will strengthen the artifacts because we have reshaped the feature from the spatial dimension into the channel dimension. In contrast, the filter-based tokenizer produces features in a much more smooth manner and outputs visual pleasant results (see Fig.~\ref{fig:spa-vis}(b)). More details can be found in the zoom-in view in Fig.~\ref{fig:spa-vis}.

\subsection{More ablation studies for each components}

\begin{figure}[t]
  \centering
  \includegraphics[width=0.6\linewidth]{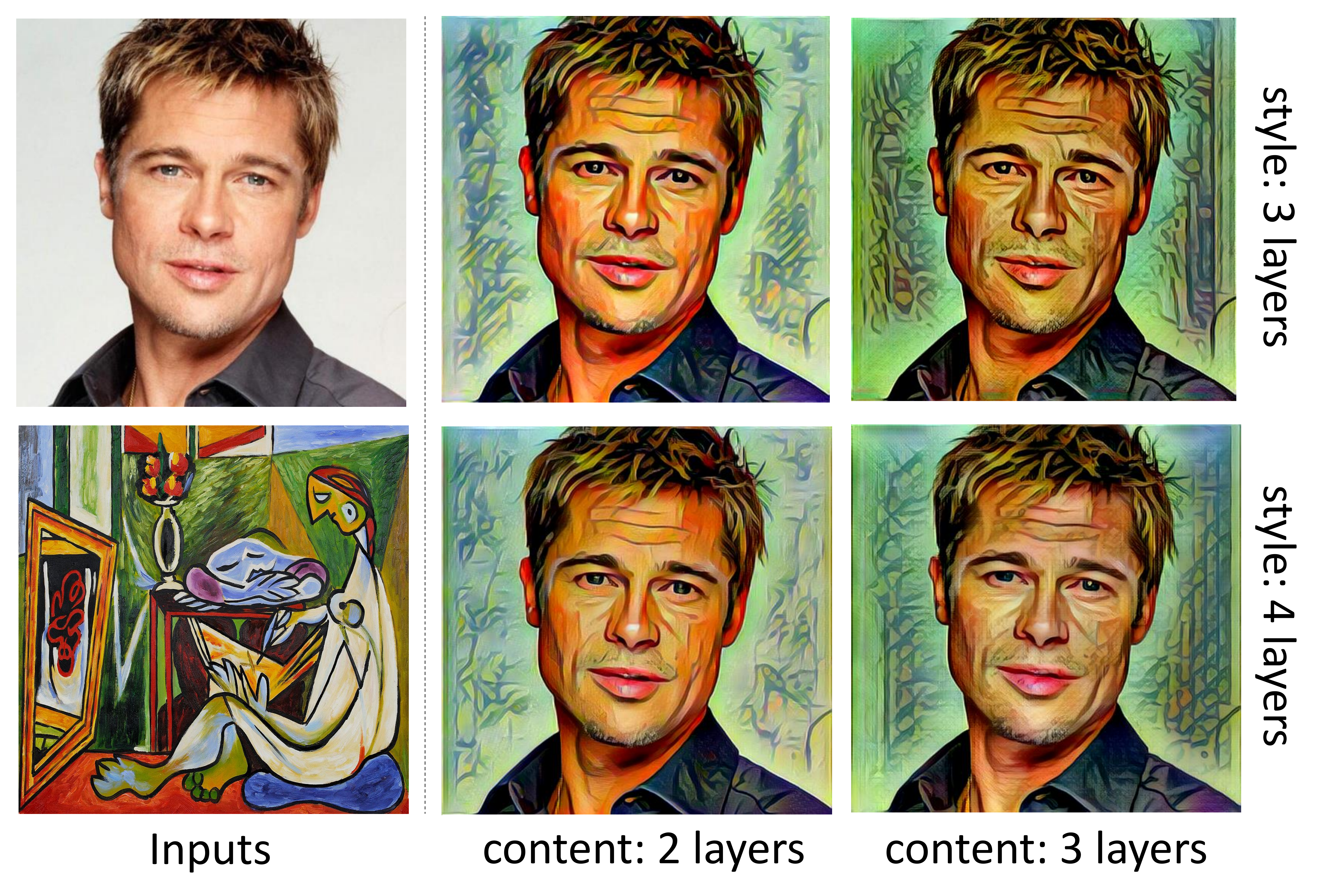}
  \caption{Visual comparison with different receptive field in the content and style backbone. 
  \label{fig:rcf}
}
\end{figure}

\subsubsection{Number of Encoder and Decoder Layers}

In Fig.~\ref{fig:enc-dec-number}, we show stylized results generated by our model with different layer numbers of encoder and decoder. We can observe model with deeper encoder and decoder has stronger capability to preserve semantic similarity, so that similar style patterns (e.g., fire) can be transferred to similar content regions. 

\subsubsection{The Receptive Field in the Backbone}\label{sec:ablation-rcfB}
In Fig.~\ref{fig:rcf}, we change the depth of content and style backbone independently. The receptive field size in the backbone could affect the stroke size. The larger receptive field for extracting content features could make the final results miss details and also spend more time converging (see the content backbone with four layers in Fig.~\ref{fig:rcf}). While for extracting style features, the larger receptive field always appears with a deeper backbone which could extract higher-level semantic features for better understanding the style pattern. For example, in Fig.~\ref{fig:rcf}, the results with deeper style backbone (4 layers) show much more smooth content. Thus, for style features, we suggest using a deeper backbone while for content features, a shallow backbone is recommended. This is also consistent with the observation in~\cite{stroke}.

In the experiments, we have also tried the Gram matrix
loss~\cite{gatys2016} to replace the AdaIN style loss~\cite{adain} as presented in
Sec. 3.2 in the main paper. However, the results show that the AdaIN style loss performs better.

\subsubsection{The Receptive Field in the loss network}\label{sec:ablation-rcfL}
\begin{figure*}[t]
  \includegraphics[width=1.0\linewidth]{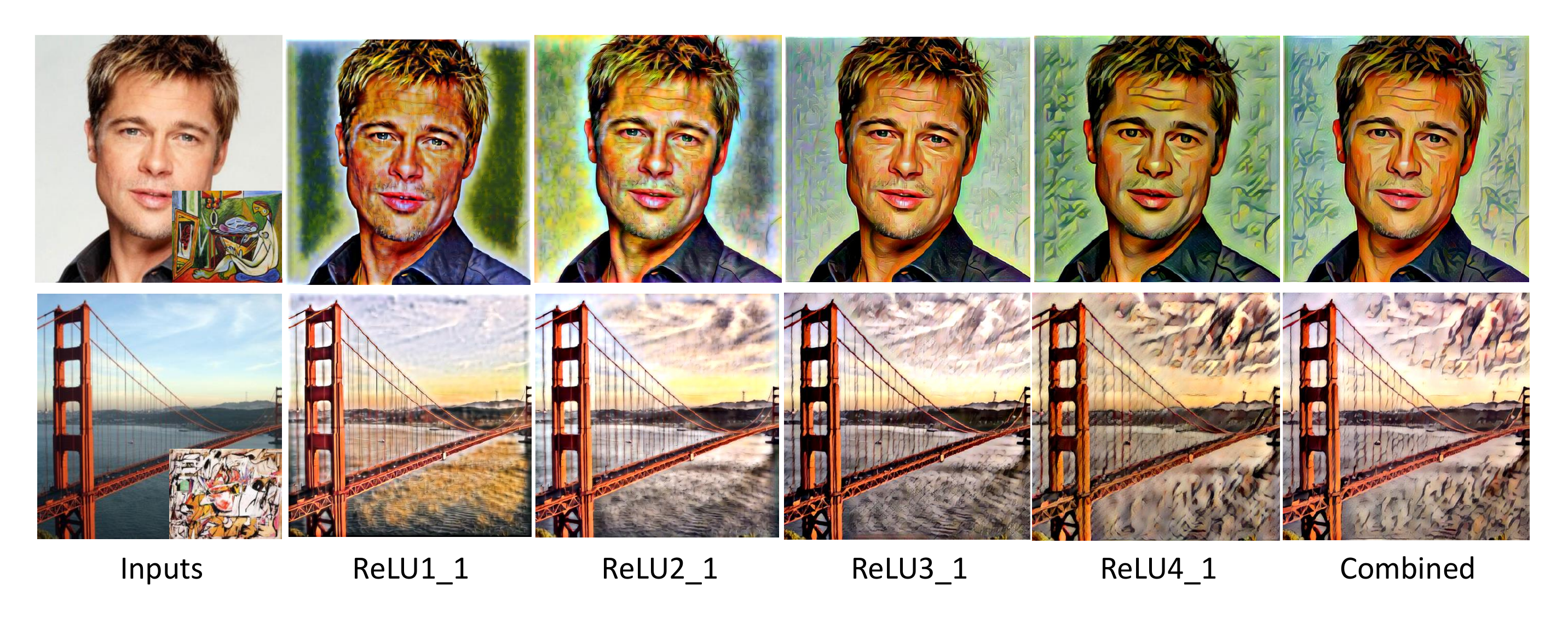}
  \caption{Results with losses at different levels. We use a fixed-weight VGG-19 as our loss network.}
  
  \label{fig:loss-depth}
\end{figure*}

Since features from different layers capture different style details. The degree of stylization can be modified by using multiple levels of features. As shown in Fig.~\ref{fig:loss-depth}, transferring content and style features from shallower layers ($relu1\_1$) produce more photo-realistic visual effects. On the other hand, using features from deeper layers bring more abstract style to the sky. Optimal results could be obtained by computing the average value of feature difference in the four layers.

\section{Control the Style Size} 
As style transfer is a very subjective task, sometimes we require realistic results (preserving more details) while sometimes we prefer artistic stylization (more abstract style). We could control the magnitude of stylization by the following three factors in our proposed STTR: 

\begin{itemize}
  \setlength{\parskip}{0cm} 
  \setlength{\itemsep}{0cm}
\item The receptive field in the loss network (
Sec. 3.2 in the main paper
and Sec.~\ref{sec:ablation-rcfL}).
\item The receptive field in the backbone (
Sec. 3.1 in the main paper
and Sec.~\ref{sec:ablation-rcfB}).
\item The loss weight $\lambda$ (
Sec. 3.2 in the main paper
and 
Sec. 4.3 in the main paper
). 
\end{itemize}

For 
Fig. 3
in the main paper, we set the content backbone with activation after 2 layers while style backbone with 4 layers, $\lambda=10$. During training and testing, increasing the receptive field of the style backbones or the loss network, or setting a larger $\lambda$, could provide richer styles.

\begin{figure*}[t]
  \includegraphics[width=1.0\linewidth]{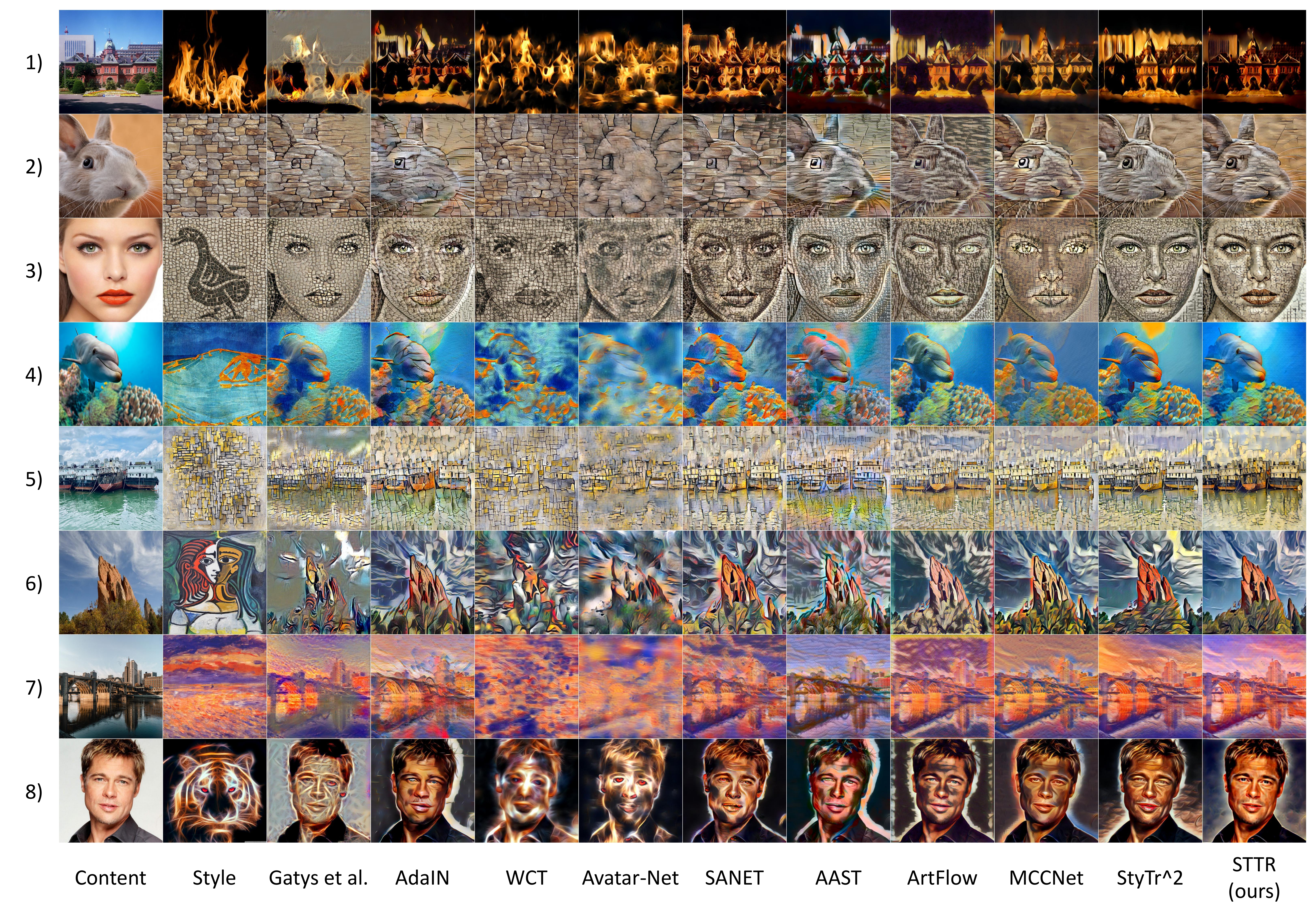}
  \caption{More Qualitative Results for Image Style Transfer}
  
  \label{fig:supp-comparison-p}
\end{figure*}

\begin{figure*}
  \centering
  \includegraphics[width=0.8\linewidth]{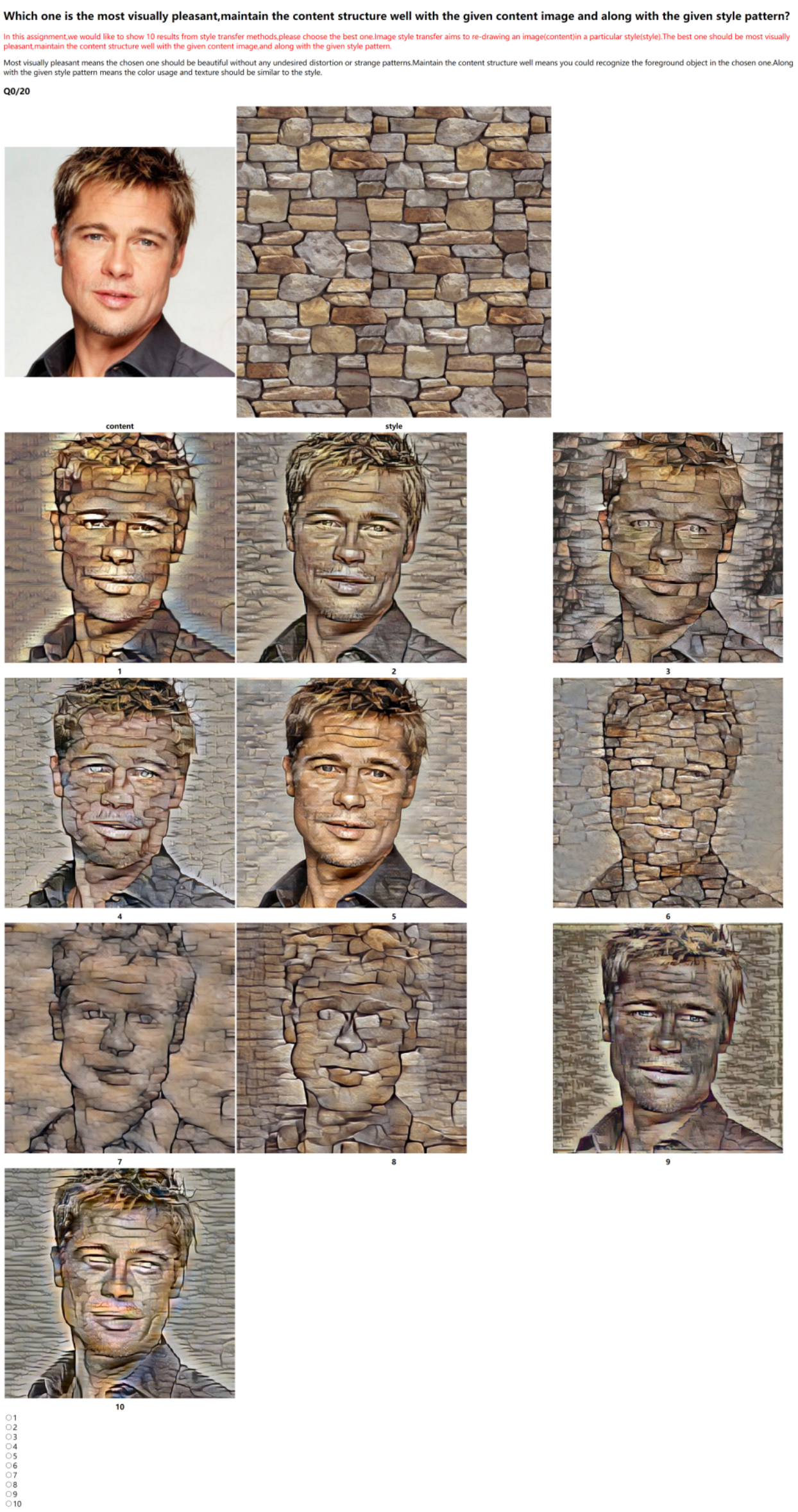}
  \caption{Questionnaire for comparison between other state-of-the-art methods in the user study. }
  
  \label{fig:supp-user1}
\end{figure*}

\section{Additional Experiments} 

\subsection{More Qualitative Results for Image Style Transfer}
We provide further comparisons obtained by our proposed STTR and other state-of-the-art methods. We evaluate various content images 
with distinctive styles. The results are illustrated in Fig.~\ref{fig:supp-comparison-p}. 
The results of compared methods are obtained by running their codes with default configurations. All of the images used in the testing stage are never observed during the training stage.

\section{Details of User Study}
To evaluate the results, we conduct a user study on Amazon Mechanical Turk (AMT). We have designed a website to show comparison results between other state-of-the-art methods and our proposed STTR. We use 10 content images and 30 style images collected from copyright-free websites. For each method, we use the released codes and default parameters to generate 300 content-style pairs. We hire 75 volunteers on Amazon Mechanical Turk (AMT) for our user study. 20 of 300 content-style pairs are randomly selected for each user. The screenshot of the designed website is shown in Fig.~\ref{fig:supp-user1}.

The instruction for this questionnaire is as follows:

\emph{Which one is the most visually pleasant, maintaining the content structure well with the given content image and along with the given style pattern?}

\emph{In this assignment, we would like to show seven results from style transfer methods, please choose the best one. Image style transfer aims to re-drawing an image (content) in a particular style (style). The best one should be most visually pleasant, maintain the content structure well with the given content image, and along with the given style pattern.}

\emph{Most visually pleasant means the chosen one should be beautiful without any undesired distortion or strange patterns. Maintaining the content structure well means you could recognize the foreground object in the chosen one. Along with the given style pattern means the color usage and texture should be similar to the style.}

Each user is asked to vote for only one result (they don't know which one is generated by which method). The average time for a volunteer to finish the questionnaire is five minutes. Finally, we collect 1500 votes from 75 users and calculate the percentage of votes that each method received.

\end{appendix}

\end{document}